\documentclass[a4paper,fleqn]{cas-dc}

\usepackage[numbers]{natbib}
\usepackage{ulem}

\usepackage{algorithm}
\usepackage{algorithmicx}
\usepackage{algpseudocode}
\usepackage{multirow}
\usepackage{amsmath}
\usepackage{booktabs}
\usepackage{graphicx}
\usepackage{fontawesome}

\usepackage{hyperref}
\hypersetup{hidelinks,
	colorlinks=true,
	allcolors=black,
	pdfstartview=Fit,
	breaklinks=true}

\def\tsc#1{\csdef{#1}{\textsc{\lowercase{#1}}\xspace}}
\tsc{WGM}
\tsc{QE}
\tsc{EP}
\tsc{PMS}
\tsc{BEC}
\tsc{DE}

\begin{document}
\let\WriteBookmarks\relax
\def\floatpagepagefraction{1}
\def\textpagefraction{.001}

\shorttitle{ActiveFreq: Integrating Active Learning and Frequency Domain Analysis for Interactive Segmentation}

\shortauthors{Lijun Guo \emph{et al.}}

\title [mode = title]{ActiveFreq: Integrating Active Learning and Frequency Domain Analysis for Interactive Segmentation}

\author[1]{Lijun Guo}
\credit{Methodology, Software, Writing - Original draft preparation}

\affiliation[1]{organization={School of Computer Science, Wuhan University},
    city={Wuhan},
    postcode={430072}, 
    country={China}}

\author[1]{Qian Zhou}
\credit{Writing - Review and editing}

\author[1]{Zidi Shi}[style=chinese]
\credit{Methodology, Software}

\author[1]{Hua Zou}
\cormark[1]
\ead{zouhua@whu.edu.cn}
\credit{Conceptualization of this study, Methodology, Writing - Review and editing, Supervision}

\author[2]{Gang Ke}
\credit{Review and editing}
\affiliation[2]{organization={Dongguan Polytechnic},
    city={Dongguan},
    postcode={523651}, 
    country={China}}

\begin{abstract}
Interactive segmentation is commonly used in medical image analysis to obtain precise, pixel-level labeling, typically involving iterative user input to correct mislabeled regions. However, existing approaches often fail to fully utilize user knowledge from interactive inputs and achieve comprehensive feature extraction. Specifically, these methods tend to treat all mislabeled regions equally, selecting them randomly for refinement without evaluating each region’s potential impact on segmentation quality. Additionally, most models rely solely on spatial domain features, overlooking frequency domain information that could enhance feature extraction and improve performance. To address these limitations, we propose ActiveFreq, a novel interactive segmentation framework that integrates active learning and frequency domain analysis to minimize human intervention while achieving high-quality labeling. ActiveFreq introduces AcSelect, an autonomous module that prioritizes the most informative mislabeled regions, ensuring maximum performance gain from each click. Moreover, we develop FreqFormer, a segmentation backbone incorporating a Fourier transform module to map features from the spatial to the frequency domain, enabling richer feature extraction. Evaluations on the ISIC-2017 and OAI-ZIB datasets demonstrate that ActiveFreq achieves high performance with reduced user interaction, achieving 3.74 NoC@90 on ISIC-2017 and 9.27 NoC@90 on OAI-ZIB, with 23.5\% and 12.8\% improvements over previous best results, respectively. Under minimal input conditions, such as two clicks, ActiveFreq reaches mIoU scores of 85.29\% and 75.76\% on ISIC-2017 and OAI-ZIB, highlighting its efficiency and accuracy in interactive medical segmentation.
\end{abstract}


\maketitle

\section{Introduction}
Deep learning-based automatic segmentation methods have achieved significant success in medical image analysis, delivering high accuracy and efficiency across various clinical applications~\cite{wang2022medical,shen2017deep, feng2024gcformer,zhao2024morestyle}. However, these methods depend on large-scale annotated datasets and complex architectures to learn patterns. This reliance creates challenges in medical image segmentation, where datasets are often limited, variable, and difficult to annotate. In contrast, interactive segmentation requires minimal user input to refine predictions and produce precise results, even in complex scenarios. With just a few clicks, users can achieve reliable, high-quality outcomes~\cite{zhao2013overview,masood2015survey}. By incorporating user expertise, interactive methods provide more accurate and robust segmentations, making them especially valuable in clinical settings where precision is essential~\cite{wang2018deepigeos}.

\begin{figure}
    \centering
    \includegraphics[scale=0.36]{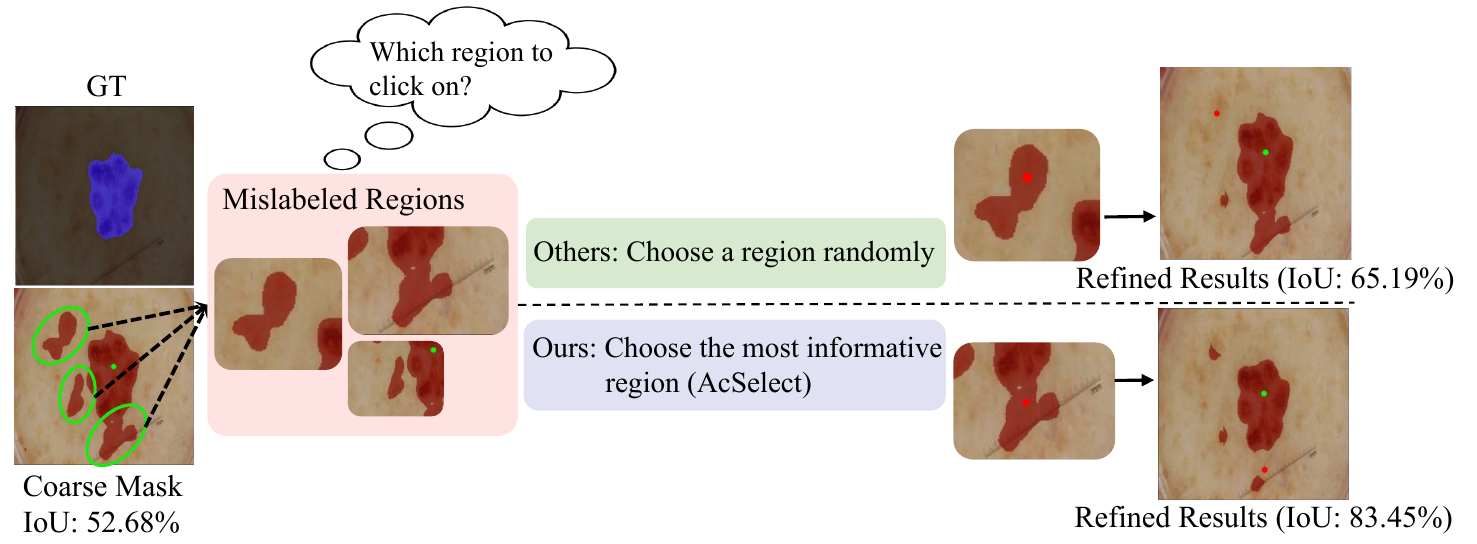}
    \caption{Comparison between our proposed method and other methods. While other methods randomly select a mislabeled region to click, our approach uses an active learning selection module (AcSelect) to evaluate all mislabeled areas and select the most valuable one for annotation.}
    \label{FIG:1}
\end{figure}

Interactive image segmentation research has explored various interaction inputs, including bounding boxes~\cite{xu2016deep, zhang2023ms}, polygons~\cite{acuna2018efficient}, clicks~\cite{sofiiuk2022reviving}, scribbles~\cite{wu2014milcut}, and their combinations~\cite{zhang2020interactive}. Among these, click-based approaches stand out due to their simplicity and well-established protocols for training and evaluation. Recent advancements in click-based segmentation have focused on two key aspects: refining segmentation outputs more efficiently and enhancing the performance of backbone networks.

To improve refinement efficiency, several methods have targeted the local adjustment of segmentation masks. Ding \emph{et al.}~\cite{ding2021dual} propose a dual-stream framework with an interactive shape stream to extract precise boundary information and refine segmentation accuracy through adaptive Gaussian maps.  FocalClick~\cite{chen2022focalclick} and FocusCut~\cite{lin2022focuscut} refine local regions based on user clicks, preserving details in unaffected areas and iteratively narrowing the refinement scope. Despite their success, these methods typically process new clicks by assigning them through the entire segmentation network, resulting in computational inefficiencies. Addressing this, EMC-Click~\cite{du2023efficient} introduces a lightweight mask correction network that updates coarse masks directly, avoiding the need to rerun the full segmentation pipeline. In parallel, significant efforts have been made to enhance backbone network performance for interactive segmentation. FCANet~\cite{lin2020interactive} incorporates first-click attention mechanisms and click-based loss functions to prioritize user inputs effectively. CGAN~\cite{yan2024cgan} leverages adversarial learning to ensure consistency between predictions and ground truths, while CIDN~\cite{zhang2024cidn} facilitates context interaction across low-level and high-level features to capture detailed structures such as vascular boundaries. These innovations have substantially improved feature representation, segmentation accuracy, and interaction efficiency.

Despite significant advancements in interactive segmentation methods, several challenges persist in scenarios involving complex anatomical structures or pathological cases. Existing methods often require numerous user interactions, with each click providing only marginal improvements in segmentation result. This inefficiency is primarily due to suboptimal handling of mislabeled regions during the refinement process. As illustrated in Fig.~\ref{FIG:1}, after generating a coarse mask and identifying multiple mislabeled regions, most existing approaches treat all regions as equally important and randomly select one for annotation. 
However, different mislabeled regions vary greatly in their potential to improve the final segmentation. To address this, we define the most informative region as the one whose correction brings the highest performance gain. It typically exhibits high prediction uncertainty, a large number of mislabeled pixels, and strong internal error consistency, where the region’s overall feature representation closely aligns with that of the most erroneous pixel.
Additionally, existing methods rely heavily on spatial domain features, overlooking the complementary advantages of frequency domain analysis. Frequency domain features are particularly beneficial in medical imaging, as they effectively filter high-frequency noise generated by imaging equipment while preserving critical low-frequency structural information~\cite{wang2024hedn, rs16193594}. This capability is essential for accurately capturing fine anatomical structures, such as blood vessels and small lesions, which are often challenging to delineate using spatial domain information alone. The absence of frequency domain analysis significantly limits the ability of current models to achieve precise segmentation.

FCANet~\cite{lin2020interactive} demonstrates that indiscriminate treatment of sequential clicks leads to suboptimal segmentation accuracy. Building on the insights of this, we observe that treating all mislabeled regions equally during refinement is similarly suboptimal. While FCANet differentiates the treatment of previously applied clicks by assigning varying weights, our approach focuses on the selective refinement of mislabeled regions that will be clicked in the future, addressing the limitation of random click region selection in FCANet.
To address these limitations, we propose \textbf{ActiveFreq}, a novel pipeline that combines an active learning selection module, \textbf{AcSelect}, and a coarse segmentation backbone, \textbf{FreqFormer}. AcSelect leverages entropy-based metrics to assess the uncertainty within each mislabeled region, assigning scores to prioritize regions with the highest uncertainty for refinement. This targeted module minimizes user interactions while maximizing segmentation accuracy. Although most active learning-based methods~\cite{lenczner2022dial, burmeister2022less, xie2022towards, siddiqui2020viewal} employ entropy-based metrics for sampling, they typically design a single metric based on individual pixels, without exploring the relationships between pixels or among specific regions. In contrast, our proposed AcSelect conducts sampling analysis from three levels: individual pixels, pixel sets, and regions, which endows it with stronger sampling capabilities. Notably, AcSelect is designed as a plug-and-play module that can be seamlessly integrated into any CNN-based or Transformer-based backbone, enhancing its versatility and adaptability across different architectures. 
The goal of AcSelect is to intelligently identify and prioritize the most informative regions from coarse segmentation masks. This helps to minimize the user interaction cost while ensuring that each click contributes maximal refinement.
\textbf{FreqFormer} enhances the segmentation backbone by improving the decoder structure of SegFormer~\cite{xie2021segformer}. Specifically, FreqFormer incorporates Fourier transform modules to extract frequency domain features, which are then fused with spatial features through a redesigned decoder. It leverages multi-dimensional frequency analysis to capture both global structural patterns and fine-grained details, serving as a robust foundation for subsequent interactive refinements.

In general, our contributions can be summarized as follows:
\begin{itemize}
    \item  We propose \textbf{ActiveFreq}, a novel interactive segmentation framework that maintains high segmentation accuracy with a few clicks. To the best of our knowledge, it is the first pipeline that prioritizes mislabeled regions based on their informativeness rather than selecting and refining them randomly.

    \item We develop \textbf{AcSelect}, a lightweight and plug-and-play active learning selection module that uses entropy-based metrics to identify and refine the most informative mislabeled regions, which helps to minimize the user interaction cost while ensuring that each click contributes maximal refinement.

    \item We introduce \textbf{FreqFormer} based on SgeFormer~\cite{xie2021segformer} that incorporates Fourier transform modules to integrate complementary spatial and frequency domain features, improving segmentation precision for complex anatomical structures.
    
    \item ActiveFreq achieves state-of-the-art performance with \textbf{3.74} NoC@90 on ISIC-2017~\cite{berseth2017isic} and \textbf{9.27} NoC@90 on OAI-ZIB~\cite{ambellan2019oaizib}, surpassing previous best results by 23.5\% and 12.8\%, respectively, while significantly reducing user clicks required for optimal IoU.
\end{itemize}

\section{Related Work}
\subsection{Mask Refinement in Interactive Segmentation}
Interactive image segmentation has received significant attention in recent years for its ability to use user input to achieve precise, pixel-level object labeling. Early methods focus on encoding user clicks as spatial information to guide segmentation. For instance, DIOS~\cite{xu2016deep} converts foreground and background clicks into distance maps, which are concatenated with the input image to provide location-specific guidance. BRS~\cite{jang2019interactive} and its extension, f-BRS~\cite{sofiiuk2020f}, frame the task as an inference-time optimization problem. BRS refines segmentation by modifying the input image or distance maps, while f-BRS optimizes intermediate layer weights to improve both speed and accuracy. Ding \emph{et al.}~\cite{ding2021dual} propose an interactive shape stream with adaptive Gaussian maps and a probability click loss function for precise boundary extraction and detail sensitivity. RITM~\cite{sofiiuk2022reviving} enhances robustness by using previous segmentation masks as input, stabilizing predictions with each click. RE-3DLVNet~\cite{ge2022re} combines 3D segmentation and reinforced quantification with a novel consistency constraint, enabling precise left ventricle volume estimation through refined volume adjustment and accurate anatomical visualization.
To further improve efficiency and accuracy, recent methods have explored lightweight architectures and advanced refinement strategies. FocalClick~\cite{chen2022focalclick} uses a progressive merging mechanism to analyze morphological relationships between previous masks and current predictions, identifying regions to update or preserve. FocusCut~\cite{lin2022focuscut} adaptively crops click-centered patches from the original image after generating a global prediction, enabling more localized refinements. SimpleClick~\cite{liu2023simpleclick} employs a non-hierarchical Vision Transformer (ViT) with Masked Auto-Encoder (MAE) pre-training, enhancing global feature extraction while reducing computational costs. EMC-Click~\cite{du2023efficient} introduces a lightweight mask correction network that updates the mask iteratively without requiring the full segmentation model to rerun after each click. Finally, CFR-ICL~\cite{sun2024cfr} proposes a Cascade-Forward Refinement strategy, allowing iterative refinement during inference without adding extra network modules.
Despite recent advances, interactive segmentation methods often treat all mislabeled regions with equal importance, reducing the effectiveness of each click and requiring more user interactions for accurate results. Additionally, these models primarily rely on spatial domain information, overlooking valuable frequency domain features that could enhance robustness by filtering out high-frequency noise and preserving essential structural details.

\subsection{Active Learning for Semantic Segmentation}
Active learning approaches for image segmentation generally fall into two main categories based on annotation strategies: image-level and region-level methods. Image-level methods select samples for annotation by comparing differences between them. For instance, Yang \emph{et al.}~\cite{yang2017suggestive} develop a framework that uses uncertainty sampling and similarity estimation to select the most representative samples. Ozdemir \emph{et al.}~\cite{ozdemir2021active} use Bayesian sample querying, combining uncertainty and representativeness in an MMD-optimized latent space to enhance annotation efficiency and segmentation performance in medical imaging. Jin \emph{et al.}~\cite{jin2022one} propose a one-shot active learning framework that integrates contrastive self-supervised learning with a diversity-based sampling strategy, significantly reducing annotation costs while maintaining high segmentation accuracy. 
Gaillochet \emph{et al.}~\cite{gaillochet2023active} introduce Stochastic Batch Selection, a novel strategy that evaluates uncertainty at the batch level rather than at the individual sample level. 
Recently, active learning research has increasingly focused on region-level approaches, which enable more precise sample selection.
Mackowiak \emph{et al.}~\cite{mackowiak2018cereals} assess various strategies, including Random, Entropy, and Vote Entropy, incorporating annotation costs measured by pixels or clicks to improve selection efficiency.
Lenczner \emph{et al.} demonstrate that although pixel-wise entropy is a common uncertainty sampling strategy in active learning, it has been shown to be more effective and concise compared to MC Dropout~\cite{gal2016dropout}, ODIN~\cite{liang2017principled}, and ConfidNet~\cite{corbiere2019addressing}. Therefore, many existing uncertainty sampling methods rely on pixel-wise entropy. For instance, DIAL~\cite{lenczner2022dial} uses the pixel entropy metric derived from softmax output to sample. Similarly, Burmeister \emph{et al.}~\cite{burmeister2022less} use pixel entropy sums to select target slices when sampling 2D slices from 3D data. However, these methods neglect which sample or slice contains the most informative regions, which limits its comprehensiveness. RIPU~\cite{xie2022towards} introduces a sample labeling method based on prediction uncertainty (PU), but still relies solely on the sum of pixel entropy throughout the sample without addressing regions requiring specific attention. ViewAL~\cite{siddiqui2020viewal} introduces View Entropy Score, which calculates the entropy of each pixel's probability distribution across categories, but it still fails to integrate the surrounding pixels for a holistic analysis. These approaches often fall short by neglecting the holistic inconsistency within the sample and employing metrics that are overly simplistic and lack comprehensiveness, failing to account for the interrelationships and complementarity between sub-regions. In contrast, our proposed AcSelect comprehensively considers three levels: individual pixels, pixel sets, and regions, evaluating from local to global perspectives, which endows it with stronger sampling capabilities.

\begin{figure*}
    \centering
    \includegraphics[scale=0.6]{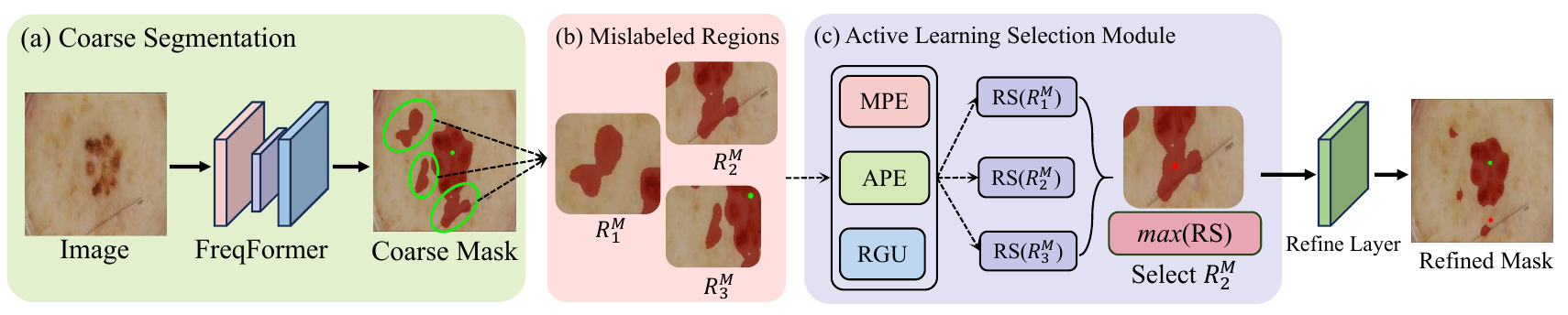}
    \caption{The overall pipeline of ActiveFreq. (a) Coarse segmentation is generated using the proposed FreqFormer. (b) Mislabeled regions within the coarse mask. (c) The AcSelect module selects the most informative region for refinement. This process evaluates each mislabeled region using three metrics: Maximum Pixel Entropy (MPE), Average Pixel Entropy (APE), and Regional Group Uncertainty (RGU). A Region Score (RS) is calculated for each region, and the region with the highest score is chosen for further refinement to produce the refined mask.}
   \label{FIG:2}
\end{figure*}

\subsection{Frequency Domain Analysis}
Recent studies have explored the integration of frequency domain analysis into deep learning, revealing its potential to enhance optimization strategies~\cite{yin2019fourier} and generalization capabilities~\cite{wang2020high}. Zhang \emph{et al.}~\cite{zhang2019making} investigate how frequency aliasing affects the shift-invariance of modern models, and subsequently, AdaBlur~\cite{zou2023delving} applies content-aware low-pass filters during downsampling for anti-aliasing. Rahaman \emph{et al.}~\cite{rahaman2019spectral} and Xu \emph{et al.}~\cite{xu2021deep} identify a phenomenon known as spectral bias (or the frequency principle), where deeper network layers prioritize learning low-frequency components during training. Similarly, Qin \emph{et al.}~\cite{qin2021fcanet} and Magid \emph{et al.}~\cite{magid2021dynamic} leverage discrete cosine transform (DCT) coefficients for channel attention mechanisms. FLC~\cite{grabinski2022frequencylowcut} highlights the adverse effects of frequency aliasing on model robustness. Recent advancements have utilized frequency components for architectural improvements. Meanwhile, Huang \emph{et al.}~\cite{huang2023adaptive} adaptive frequency filters, inspired by the convolution theorem, to act as global token mixers in deep networks. Such approaches enable models to capture non-local features more effectively~\cite{huang2023adaptive, chi2020fast, rao2021global, li2020fourier, guibas2021adaptive}. Beyond architectural improvements, frequency domain analysis addresses practical challenges in computer vision. Chen \emph{et al.}~\cite{chen2023instance} demonstrate that low-pass filters can suppress high-frequency noise in low-light images, significantly improving instance segmentation performance. Moreover, adversarial attack studies~\cite{yin2019fourier, luo2022frequency,  jia2022exploring} reveal that manipulating high-frequency components severely degrades feature representations. Luo \emph{et al.}~\cite{luo2022frequency} show that perturbing high frequencies reduces intra-category similarity, thereby undermining model robustness. These studies demonstrate the effectiveness of frequency domain analysis in various computer vision tasks, motivating us to integrate it into interactive medical image segmentation for enhanced accuracy and efficiency.

\section{Method}
In this section, we present the proposed pipeline ActiveFreq, as illustrated in Fig.~\ref{FIG:2}. In Sec.\hyperlink{sec3.1}{3.1}, we introduce AcSelect, a module for identifying the most informative mislabeled region for annotation, based on three key metrics: maximum pixel entropy (MPE), average pixel entropy (APE), and regional group uncertainty (RGU). In Sec.\hyperlink{sec3.2}{3.2}, we describe FreqFormer in detail, which incorporates a multi-dimensional Fourier transform module into the SegFormer~\cite{xie2021segformer} architecture, enabling the network to capture frequency domain information effectively.

To identify all mislabeled regions, we define mislabeled pixels by comparing the initial prediction with the ground truth (GT). If a pixel’s prediction matches the GT, it is considered correct (true), while a mismatch is considered a mislabeled pixel (false). We then employ the concept of connected components, grouping together all contiguous mislabeled pixels. This process involves identifying all connected regions of mislabeled pixels, where each region is defined as a set of pixels that are connected to each other based on a defined connectivity criterion. Once these connected components are identified, all subsequent operations on the regions are based on these connected subgraphs, enabling targeted refinement and correction of the segmentation results.

\hypertarget{sec3.1}{
\subsection{AcSelect using Active Learning}
AcSelect is designed to identify the most informative region for annotation by assessing the uncertainty across all mislabeled regions in the coarse segmentation mask. 
Informative regions are those where correction is likely to yield the greatest performance gain. Specifically, such a region typically satisfies three criteria: (1) high prediction uncertainty, reflecting the model’s low confidence in that area; (2) a large number of mislabeled pixels, indicating a significant concentration of errors; and (3) high error consistency, where the region’s overall feature representation closely aligns with that of the most erroneous pixel, suggesting that the whole region is consistently mislabeled and suitable for effective correction. These properties collectively define the most informative region, which offers the greatest corrective value per interaction.
According to these, AcSelect uses three metrics based on pixel entropy: maximum pixel entropy (MPE), average pixel entropy (APE), and regional group uncertainty (RGU). Pixel entropy assesses the model’s confidence by examining the probability distribution across classes for each pixel~\cite{Sengupta_2020_efficientEntropy, Jeon2021entropy_algorithms}. High entropy values indicate similar predicted probabilities across classes, which reflects high uncertainty and often highlights areas where the model struggles, making these metrics effective for identifying mislabeled regions~\cite{Sparavigna2019entropy}. By leveraging MPE, APE, and RGU, AcSelect prioritizes the most uncertain regions for refinement, improving segmentation accuracy with each refinement iteration.
}

\subsubsection{Maximum Pixel Entropy}
The maximum pixel entropy (MPE) score estimates pixel-level uncertainty by identifying the highest entropy value within a mislabeled region. This metric reflects the model’s confidence in its prediction for the most uncertain pixel, effectively highlighting regions containing pixels with high uncertainty. MPE is defined as:
\begin{equation}
    {\rm MPE}(R^M) = \max_{i \in R^M} [ -\sum_{c=0}^{C} p(y_i = c) \log p(y_i = c) ]
\end{equation}
where $R^M$ denotes a mislabeled region, and $p(y_i = c)$ is the predicted probability of pixel $i$ belonging to class $c$.

\subsubsection{Average Pixel Entropy}
While MPE targets the most uncertain pixel, the average pixel entropy (APE) score assesses the overall uncertainty across all pixels within a mislabeled region. APE provides a more holistic measure of prediction uncertainty by averaging the entropy values of all pixels in the region, defined as:
\begin{equation}
    {\rm APE}(R^M) = -\frac{1}{|R^M|} \sum_{i \in R^M} \sum_{c=0}^{C} p(y_i = c) \log p(y_i = c)
\end{equation}
where $|R^M|$ represents the number of pixels in the mislabeled region $R^M$.

\begin{figure*}
    \centering
    \includegraphics[scale=0.475]{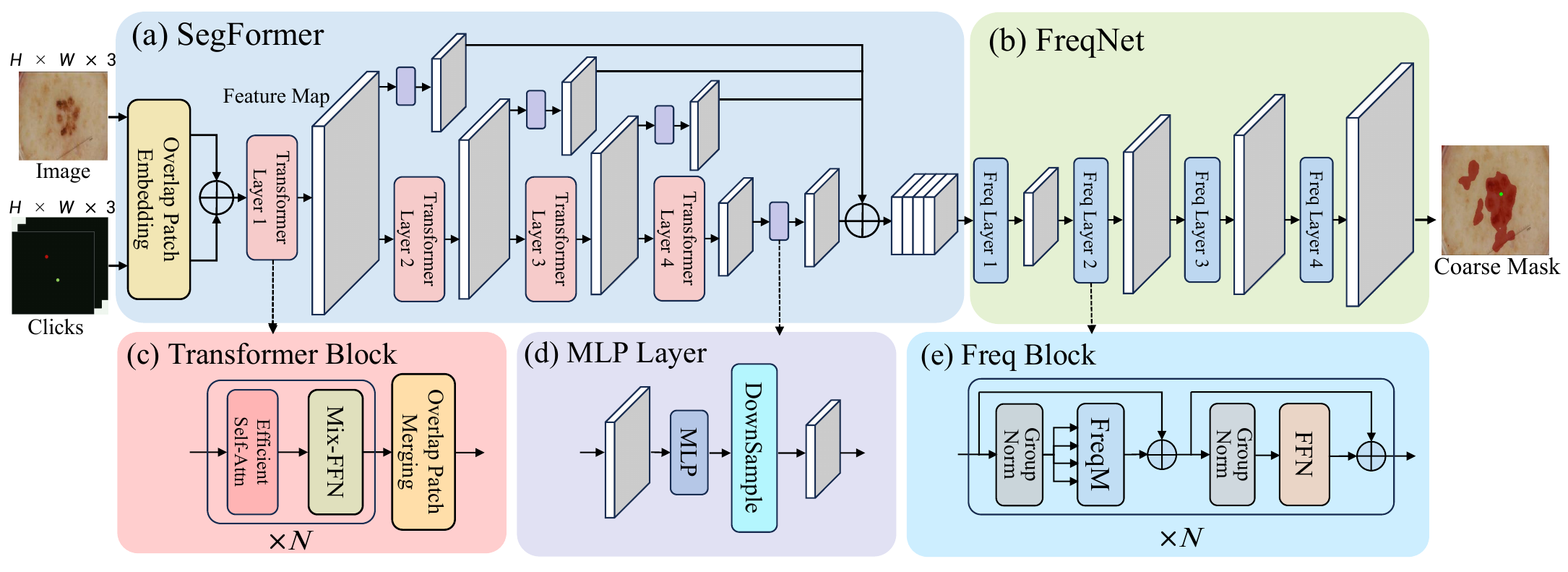}
    \caption{The diagram of FreqFormer. (a) SegFormer encoder extracts multi-scale features from the input image and user clicks. (b) FreqNet decoder, composed of four Freq layers, refines these features to produce a coarse mask. (c) Transformer Block architecture, where FFN means feed-forward network. (d) MLP Layer aligns feature maps for final concatenation. (e) Freq Block architecture in the Freq layer, in which FreqM represents the proposed frequency analysis module.}
    \label{FIG:3}
\end{figure*}

\subsubsection{Regional Group Uncertainty}
The regional group uncertainty (RGU) score estimates uncertainty at the region level by leveraging the assumption that pixels within the same region tend to share similar characteristics, such as color or texture~\cite{HARALICK1985100, tripathi2012image}. This metric identifies a dominant pixel within a mislabeled region, which exhibits the highest degree of error, and compares it to the average predicted probability of the region. The difference between this dominant pixel and the regional average reflects the inconsistency within the region, providing a measure of uncertainty. RGU is calculated as:
\begin{equation}
    {\rm RGU}(R^M) = 1 - \log |P_e - \overline{P}|
\end{equation}
\begin{equation}
    P_e = 
        \begin{cases} 
            \max(P_i), & R^M \in \textit{FP} \\
            \min(P_i), & R^M \in \textit{FN} 
        \end{cases}
\end{equation}
\begin{equation}
    \overline{P} = \frac{1}{|R^M|} \sum_{i \in R^M} \sum_{c=0}^{C} p(y_i = c)
\end{equation}
where $P_e$ is the predicted probability of the pixel with the highest degree of error, and $\overline{P}$ is the average predicted probability of all pixels within $R^M$. $|R^M|$ represents the number of pixels in the mislabeled region $R^M$. \textit{FP} and \textit{FN} denote false positive and false negative regions, respectively.

\subsubsection{Mislabeled Region Selection}
The selection of mislabeled regions is performed in two steps. First, we calculate the above three metrics for all mislabeled regions in the initial prediction. These scores are then combined using weighted coefficients to ensure balanced contributions from each metric, defining the overall region score (RS) as follows:
\begin{equation}
    {\rm RS}(R^M) = w_a \cdot {\rm MPE} + w_b \cdot {\rm APE} + w_c \cdot {\rm RGU}
\end{equation}

In the second step, we select the region with the highest region score, which represents the area with the greatest segmentation uncertainty. This region is prioritized for refinement, as it offers the highest potential for improving segmentation accuracy through targeted manual intervention. The selection process is formalized as:
\begin{equation}
    R_{selected} = \text{argmax}({\rm RS}(R^M))
\end{equation}

After selecting the target mislabeled region and performing interactive annotation, the sample with the new click is fed into the refinement layer for further processing.

\subsection{Network Architecture of FreqFormer}
\hypertarget{sec3.2}{
This section introduces the proposed network architecture, FreqFormer, as illustrated in Fig.~\ref{FIG:3}. FreqFormer consists of two main components: the SegFormer encoder~\cite{xie2021segformer} and the FreqNet decoder. SegFormer combines a Vision Transformer (ViT)~\cite{alexey2020image} backbone with a lightweight MLP decoder for efficient feature extraction and transformation. The proposed FreqNet enhances segmentation by integrating complementary frequency domain information into spatial domain features.
}

\subsubsection{SegFormer Encoder}
As shown in Fig.~\ref{FIG:3} (a), the SegFormer encodes the input image and the click map into a concatenated feature map for further processing. It consists of four hierarchical transformer layers that use self-attention to capture multi-resolution features. Four lightweight MLP layers unify the feature maps from each stage into a consistent resolution, preparing them for concatenation and subsequent decoding.

\subsubsection{FreqNet Decoder}
The FreqNet decoder, shown in Fig.~\ref{FIG:3} (b), is designed to address limitations in existing methods that primarily focus on spatial domain information. Spatial domain features often struggle to distinguish boundaries between objects and backgrounds, while frequency domain features provide complementary information by separating objects based on their frequency components~\cite{huang2021medical}. The FreqNet consists of four Freq layers and each layer comprising \(N\) Freq blocks, with each block progressively refining the features extracted by the SegFormer backbone. As depicted in Fig.\ref{FIG:3} (e), each Freq block is composed of a sequence of operations: Group Normalization (GN), Frequency Analysis Module (FeqModule), another Group Normalization, and a Feed-Forward Network (FFN). Residual connections are applied after the FreqModule and FFN, allowing the network to retain information from previous layers while refining the features.

\subsubsection{Frequency Analysis Module}
\begin{figure}
    \centering
    \includegraphics[scale=0.24]{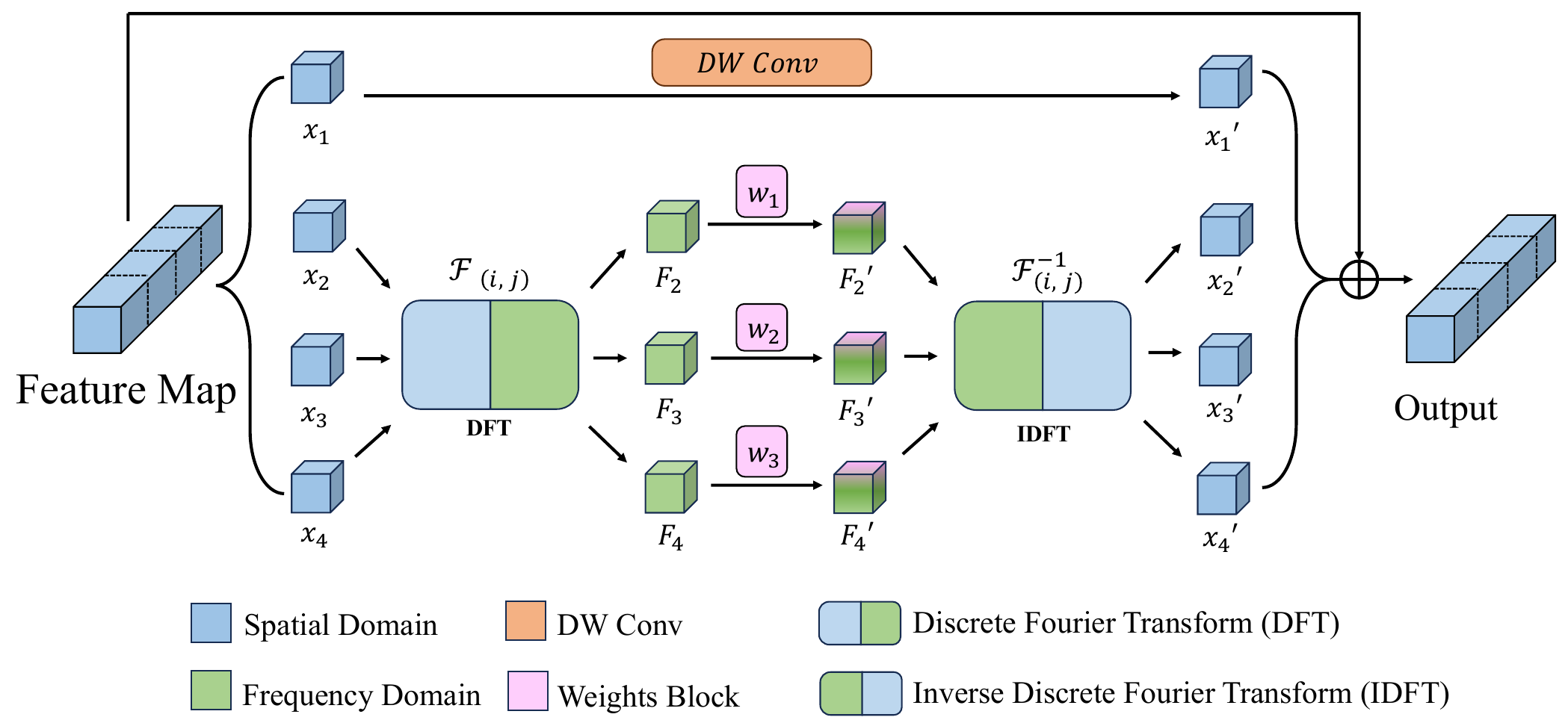}
    \caption{The architecture of frequency analysis module (FreqModule). The input feature map is split into four sub-feature maps: one is processed with depth-wise convolution (DW Conv) for spatial information, while the other three pass through 2D Discrete Fourier Transform (DFT) and 2D Inverse DFT (IDFT) for frequency information. The outputs are then concatenated to produce the final result.}
    \label{FIG:4}
\end{figure}

Fig.~\ref{FIG:4} shows the details of the Frequency Analysis Module (FreqModule). It processes feature maps to extract both spatial and frequency domain information. Given an input feature map of dimensions $H \times W \times C$, the channel dimension is divided uniformly into four parts, producing four sub-feature maps $\{x_1, x_2, x_3, x_4\}$ of size $H \times W \times \frac{C}{4}$. More specifically, the first sub-feature map is processed using a depth-wise convolution (DWConv) to retain spatial domain information. The remaining three sub-feature maps are sequentially processed using 2D Discrete Fourier Transform (2D DFT) and 2D Inverse Discrete Fourier Transform (2D IDFT), extracting frequency domain features. These operations allow the network to capture complementary information that aids in distinguishing object boundaries.

\noindent \textbf{Depth-wise Separable Convolution (DW Conv).} 
The first sub-feature map, denoted as $x_1$, is processed using depth-wise separable convolution to retain spatial features. This operation is expressed as: 
\begin{equation} 
    x'_1 = {\rm DW Conv}(x_1) 
\end{equation}

\noindent \textbf{2D Discrete Fourier Transform (2D DFT).} 
The remaining three sub-feature maps, $x_2$, $x_3$, and $x_4$, are transformed into the frequency domain using 2D DFT to capture spatial and cross-channel frequency information. Each sub-feature map undergoes 2D DFT along different dimension pairs, \emph{i.e.}, along the height $H$ and width $W$ for $x_2$, along the height $H$ and channel $C$ for $x_3$, and along the width $W$ and channel $C$ for $x_4$:
\begin{equation}
    F_2(u, v, c) = \sum ^{H-1}_{x=0} {\sum ^{W-1}_{y=0} {{x}_{2}}}(x,\,y,\,c)\cdot{e}^{-j2\pi(\frac {ux} {H}+\frac {vy} {W})}
\end{equation}
\begin{equation}
    F_3(u, y, v) = \sum ^{H-1}_{x=0} {\sum ^{\frac{C}{4}-1}_{c=0} {{x}_{3}}}(x,\,y,\,c)\cdot{e}^{-j2\pi(\frac {ux} {H}+\frac {vc} {C/4})}
\end{equation}
\begin{equation}
    F_4(x, u, v) = \sum ^{W-1}_{y=0} {\sum ^{\frac{C}{4}-1}_{c=0} {{x}_{4}}}(x,\,y,\,c)\cdot{e}^{-j2\pi(\frac {uy} {W}+\frac {vc} {C/4})}
\end{equation}
where $(x,\,y,\,c)$ is the coordinates in the spatial domain $(u,\,v,\,c)$, $(u,\,y,\,v)$ and $(x, u, v)$ are the coordinates in the frequency domain. $j$ is an imaginary unit.

\noindent \textbf{2D Inverse Discrete Fourier Transform (2D IDFT).} 
Each frequency-transformed sub-feature map is assigned a unique weight $w_i$, applied to the corresponding 2D DFT output to adjust its contribution. We then perform 2D IDFT to convert these frequency domain features back to the spatial domain as follows:
\begin{equation}
    F'_i = w_i \cdot F_i, \quad i = 2, 3, 4
\end{equation}
\begin{equation}
    x'_i = \mathcal{F}_{\text{2D}}^{-1}(F'_i), \quad i = 2, 3, 4
\end{equation} 
where $w_i$ represents the unique weights for each feature map, and $\mathcal{F}_{\text{2D}}^{-1}$ denotes the 2D IDFT operation. On the $(H, W)$ dimensions, the 2D IDFT is given by: 
\begin{equation} 
    f(x, y, c) = \frac{1}{H \cdot W} \sum_{u=0}^{H-1} \sum_{v=0}^{W-1} F_1(u, v, c) \cdot e^{j 2 \pi \left( \frac{ux}{H} + \frac{vy}{W} \right)} 
\end{equation}
For 2D IDFT transformations along $(H, C)$ dimensions and $(W, C)$ dimensions, a similar operation is used with corresponding summation indices.

Finally, the processed sub-feature maps $x'_1$, $x'_2$, $x'_3$, and $x'_4$ are concatenated along the channel dimension to form the output feature map $x'$, matching the shape of the original input feature map and serving as the output of the FreqModule.

\subsection{Refinement Process}
In the refinement process, following FocalClick~\cite{chen2022focalclick}, we employ a series of methods to ensure that previously correctly segmented regions remain unaffected. First, we assign smaller weights to correctly segmented regions or completely exclude them from updates, reducing their influence during refinement. This allows the model to focus on correcting the areas that need attention. Additionally, we leverage ROI Align~\cite{he2017mask}, which precisely extract features from regions of interest (ROI) in the feature map, ensuring that spatial information is preserved without any quantization errors. By using bilinear interpolation instead of fixed grid pooling, ROI Align accurately aligns the features with the original object boundaries. This method guarantees that the correctly segmented regions remain unaffected during refinement, as it selectively focuses on the areas requiring adjustment, without altering the previously accurate segmentation, which maintains the integrity of correctly segmented areas.

The Refine Layer is the last module in the refinement process, which decodes the output of FreqFormer and maps it to the original size. Following FocalClick~\cite{chen2022focalclick}, the specific structure is as follows:
\begin{equation}
    {\rm output} = {\rm XConvBlock(ConvBlock}(x'))
\end{equation}
where $x'$ represents the output of the FreqFormer. ConvBlock consists of a convolutional layer followed by batch normalization and a ReLU activation function (Conv + BN + ReLU). XConvBlock is composed of a 3$\times$3 convolutional layer, a 1$\times$1 convolutional layer, batch normalization, and a ReLU activation function (3$\times$3 Conv + 1$\times$1 Conv + BN + ReLU).

\section{Experiments}
To show the effectiveness of ActiveFreq, we compare our method with seven state-of-the-art interactive segmentation approaches, including CDNet~\cite{cdnet}, RITM~\cite{sofiiuk2022reviving}, iSegFormer~\cite{liu2022isegformer}, FocalClick~\cite{chen2022focalclick}, SimpleClick~\cite{liu2023simpleclick}, GPCIS~\cite{zhou2023interactive}, and GraCo~\cite{Zhao_2024_CVPR}. For a more comprehensive comparison, we also introduce two variants of our model to validate the flexibility and adaptability of the proposed AcSelect module. In Variant-1, AcSelect is directly applied to FocalClick~\cite{chen2022focalclick} based on HRNet-18~\cite{sun2019high}, while Variant-2 applies AcSelect to FocalClick with SegFormerB0~\cite{xie2021segformer} as the backbone.

\subsection{Experimental setups}
\noindent \textbf{Datasets}. 
We conduct experiments on two datasets: ISIC-2017~\cite{berseth2017isic} and OAI-ZIB~\cite{ambellan2019oaizib}. ISIC-2017 is a large-scale dataset of dermoscopy images from the International Skin Imaging Collaboration (ISIC), containing 2000 images for training, 600 for validation, and 150 for testing. The OAI-ZIB dataset consists of 507 3D MRI images with annotations for the femur, tibia, patella, and tibial cartilage. Following iSegFormer~\cite{liu2022isegformer}, we focus on cartilage segmentation, selecting three specific slices (slices 40, 80, and 120) from each image, resulting in 1521 training slices, 150 validation slices, and 150 testing slices.

\noindent \textbf{Evaluation Metrics.} 
Consistent with previous studies~\cite{liu2023simpleclick, sofiiuk2022reviving}, we evaluate performance using the Number of Clicks (NoC) required to reach predefined Intersection-over-Union (IoU) thresholds of 80\%, 85\% and 90\%, denoted as NoC@80, NoC@85, and NoC@90, respectively. The maximum number of clicks per instance is capped at 20. Additionally, following SimpleClick~\cite{liu2023simpleclick}, we report the mean IoU at a fixed number of clicks (mIoU@$k$) to assess segmentation quality under limited user input.

\begin{table}[htbp]
    \caption{Configuration of the FreqFormer architecture, including transformer layers in the encoder and Freq layers in the decoder. Embed dim means the channel $C$ of feature maps, and attn heads are the number of heads in self-attention.}
    \label{table4}
    \centering
    \renewcommand\arraystretch{1.1}
    \resizebox{0.5\textwidth}{!}{%
    \begin{tabular}{l|cccc}
        \Xhline{1px}
        Layer Type & Layer Index & Embed Dim & Attn Heads & Block Num \\ 
        \hline
        \multirow{4}{*}{Transformer} & 1 & 32 & 1 & 2     \\
                                           & 2 & 64 & 2 & 2     \\
                                           & 3 & 128 & 4 & 2     \\
                                           & 4 & 256 & 8 & 2     \\ 
        \hline
        \multirow{4}{*}{FreqLayer} & 1 & 256 & - & 1     \\
                                    & 2 & 128 & - & 1     \\
                                    & 3 & 64 & - & 1     \\
                                    & 4 & 32 & - & 1     \\
        \Xhline{1px}
    \end{tabular}%
    }
\end{table}

\noindent \textbf{Implementation Details.} 
The encoder of our FreqFormer mainly consists of four transformer layers while the decoder contains four Freq layers. The detailed configurations are presented in Tab.~\ref{table4}. Input images are resized to $224 \times 224$ for ISIC-2017~\cite{berseth2017isic} and $384 \times 384$ for OAI-ZIB~\cite{ambellan2019oaizib}. For the Region Score (RS) computation, the weights are set as follows: $w_a=0.35$, $w_b=0.35$, and $w_c=0.30$.
We train all models for 120 epochs, with an initial learning rate of 5e-5, decaying to 5e-6 after 50 epochs. Batch size is set to 16 for FreqFormer. All experiments are conducted on two NVIDIA RTX 4090 GPUs using Pytorch. Data augmentation techniques are employed to avoid overfitting, including random resizing (0.75 to 1.25 scale), flipping, rotation, brightness adjustment, and cropping. The Adam~\cite{kingma2014adam} optimizer with $\beta_1 = 0.9$ and $\beta_2 = 0.999$ is adopted during training.

\begin{table*}[!ht]
    \caption{Quantitative comparison of NoC performance with other methods. Results are reported on two medical benchmarks: ISIC-2017~\cite{berseth2017isic} and OAI-ZIB~\cite{ambellan2019oaizib}. The best scores are highlighted in bold, and the second-best are underlined.}
    \label{table1}
    \centering
    \renewcommand\arraystretch{1.2}
    \resizebox{\textwidth}{!}{
        \begin{tabular}{lllcccccc}
            \toprule[1pt]
            \multicolumn{1}{c}{} & \multicolumn{1}{c}{} & \multicolumn{1}{c}{} & \multicolumn{3}{c}{ISIC-2017~\cite{berseth2017isic}} & \multicolumn{3}{c}{OAI-ZIB~\cite{ambellan2019oaizib}} \\
            \multirow{-2}{*}{Method} & \multirow{-2}{*}{Venue} & \multirow{-2}{*}{Backbone} & NoC@80 & NoC@85 & NoC@90 & NoC@80 & NoC@85 & NoC@90 \\ 
            \hline
            CDNet~\cite{cdnet} & ICCV2021 & ResNet-34 & 4.18 & 5.18 & 6.79 & 6.20 & 8.56 & 9.61 \\
            RITM~\cite{sofiiuk2022reviving} & ICIP2022 & HRNet-18 & 2.90 & 3.70 & 5.69 & 8.41 & 14.53 & 19.25 \\
            RITM~\cite{sofiiuk2022reviving} & ICIP2022  & HRNet-32 & 5.99 & 7.77 & 11.28 & 8.93 & 14.84 & 19.56 \\
            iSegFormer~\cite{liu2022isegformer} & MICCAI2022 & HRNet-32 & 2.73 & 3.66 & 5.91 & 8.37 & 12.99 & 18.81 \\
            FocalClick~\cite{chen2022focalclick} & CVPR2022 & HRNet-18 & 2.83 & 3.63 & \underline{5.01} & 5.75 & 8.17 & 9.83 \\
            FocalClick~\cite{chen2022focalclick} & CVPR2022 & HRNet-32 & 4.20 & 5.52 & 8.34 & 6.11 & 7.92 & \underline{9.39} \\ 
            GPCIS~\cite{zhou2023interactive} & CVPR2023 & ResNet-50 & \textbf{2.24} & \underline{3.23} & 5.52 & \underline{4.84} & \textbf{7.20} & 9.69 \\
            Variant-1 (Ours) & - & HRNet-18 & \underline{2.27} & \textbf{2.86} & \textbf{4.08} & \textbf{4.75} & \underline{7.34} & \textbf{9.35} \\ 
            \hline
            iSegFormer ~\cite{liu2022isegformer} & MICCAI2022 & Swin-B & 2.91 & 3.93 & 6.61 & 16.37 & 18.41 & 19.88 \\
            iSegFormer~\cite{liu2022isegformer} & MICCAI2022 & Swin-L & 10.03 & 11.56 & 13.96 & 19.94 & 20+ & 20+ \\
            FocalClick~\cite{chen2022focalclick} & CVPR2022 & SegFormerB0 & 2.27 & 3.05 & 5.13 & 5.54 & 8.35 & 9.66 \\
            SimpleClick~\cite{liu2023simpleclick} & ICCV2023 & ViT-B & 2.32 & 2.98 & 4.89 & 7.68 & 13.23 & 19.29 \\
            SimpleClick~\cite{liu2023simpleclick} & ICCV2023 & ViT-L & 2.86 & 3.71 & 5.89 & 7.34 & 13.23 & 19.30 \\
            GPCIS~\cite{zhou2023interactive} & CVPR2023  & SegFormerB0 & 2.18 & 2.95 & 4.29 & 5.23 & \underline{7.67} & 9.66 \\
            GraCo~\cite{Zhao_2024_CVPR} & CVPR2024 & ViT-B & 2.23 & 2.84 & 4.16 & 5.59 & 7.82 & \underline{9.41} \\
            GraCo~\cite{Zhao_2024_CVPR} & CVPR2024 & ViT-L & 2.31 & 2.97 & 4.34 & 5.25 & 7.79 & 9.42 \\ 
            Variant-2 (Ours) & - & SegFormerB0 & \underline{1.90} & \underline{2.52} & \underline{4.09} & \underline{4.58} & 7.71 & 9.64 \\ 
            ActiveFreq (Ours) & - & FreqFormer & \textbf{1.74} & \textbf{2.21} & \textbf{3.74} & \textbf{4.57} & \textbf{6.91} & \textbf{9.27} \\ 
            \bottomrule[1pt]
        \end{tabular}
    }
\end{table*}

\subsection{Comparison with state-of-the-art approaches}
\subsubsection{Quantitative Analysis}
\noindent \textbf{NoC Performance.} As shown in Tab.~\ref{table1}, ActiveFreq establishes new state-of-the-art results on both the ISIC-2017~\cite{berseth2017isic} and OAI-ZIB~\cite{ambellan2019oaizib} datasets. Specifically, it achieves 3.74 NoC@90 on ISIC-2017, surpassing the previous best method, GraCo-ViT-B~\cite{Zhao_2024_CVPR} (4.16 NoC@90) by 10.1\%, and 9.27 NoC@90 on OAI-ZIB, improving upon FocalClick-HRNet-32~\cite{chen2022focalclick} (9.39 NoC@90) by 0.12 clicks. These results demonstrate ActiveFreq’s capability to reduce user interactions while maintaining superior segmentation accuracy. The superiority of ActiveFreq stems from AcSelect’s ability to prioritize informative mislabeled regions and FreqFormer’s integration of spatial and frequency domain features for enhanced feature representation.

\begin{figure*}[htbp]
	\centering
	\includegraphics[scale=0.35]{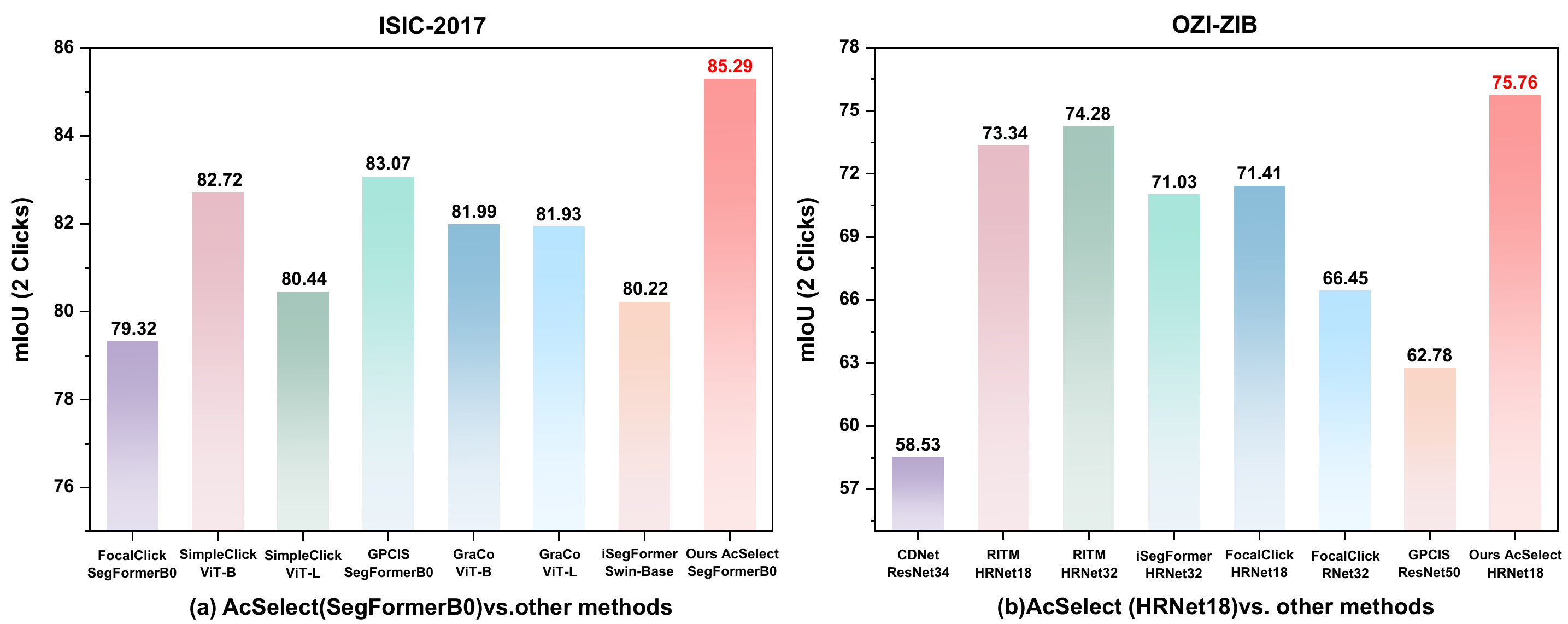}
	\caption{Comparison of mIoU@2 performance between our proposed AcSelect and other state-of-the-art methods on two datasets: (a) Transformer-based approaches on ISIC-2017~\cite{berseth2017isic} and (b) CNN-based methods on OAI-ZIB~\cite{ambellan2019oaizib}.}
	\label{FIG:5}
\end{figure*}

The effectiveness of AcSelect is further validated by the performance of the two model variants. Variant-1 (AcSelect-HRNet18) achieves 4.08 NoC@90 on ISIC-2017 and 9.35 NoC@90 on OAI-ZIB, compared to 5.01 NoC@90 and 9.83 NoC@90 for the original FocalClick-HRNet18~\cite{chen2022focalclick}, highlighting a substantial improvement.
Similarly, Variant-2 (AcSelect-SegFormerB0) achieves 4.09 NoC@90 on ISIC-2017 and 9.64 NoC@90 on OAI-ZIB, outperforming the original FocalClick-SegFormerB0~\cite{chen2022focalclick}, which records 5.13 NoC@90 and 9.66 NoC@90, respectively.
These results confirm the adaptability and robustness of AcSelect across CNN-based and Transformer-based backbones, demonstrating its capability to optimize the refinement process and enhance segmentation accuracy.

The performance gap between ActiveFreq and its variants (Variant-1 and Variant-2) underscores the additional advantages introduced by FreqFormer. On the ISIC-2017 dataset, ActiveFreq achieves 2.21 NoC@85, demonstrating a 22.73\% and 12.31\% reduction in the number of clicks required to reach the convergence threshold compared to Variant-1 (AcSelect+HRNet18) and Variant-2 (AcSelect+SegFormerB0), respectively. Similarly, on the OAI-ZIB dataset, ActiveFreq achieves 6.91 NoC@85, achieving a 5.86\% and 10.38\% reduction in click requirements compared to Variant-1 and Variant-2, respectively. These improvements highlight the effectiveness of FreqFormer in integrating spatial and frequency domain features, enabling more comprehensive feature extraction and enhanced segmentation accuracy, particularly in challenging scenarios involving complex anatomical structures or low-contrast regions.

\noindent \textbf{User Intervention.}
To evaluate the effectiveness of AcSelect under limited user interactions, Fig.\ref{FIG:5} illustrates the mIoU@2 scores on the ISIC-2017\cite{berseth2017isic} and OAI-ZIB~\cite{ambellan2019oaizib} datasets. On ISIC-2017 (Fig.\ref{FIG:5}(a)), variant-2 (AcSelect-SegFormerB0) achieves the highest mIoU of 85.29\%, surpassing the second-best method, GPCIS-SegFormerB0~\cite{zhou2023interactive} (83.07\%), by 2.22\%. Similarly, on OAI-ZIB (Fig.\ref{FIG:5}(b)), variant-1 reaches an mIoU@2 of 75.76\%, outperforming the next-best method, RITM-HRNet-32~\cite{sofiiuk2022reviving} (74.28\%), by 1.48\%. These results underscore the efficiency of AcSelect in achieving superior segmentation accuracy with minimal user input.

\begin{figure*}[htbp]
	\centering
		\includegraphics[scale=0.3]{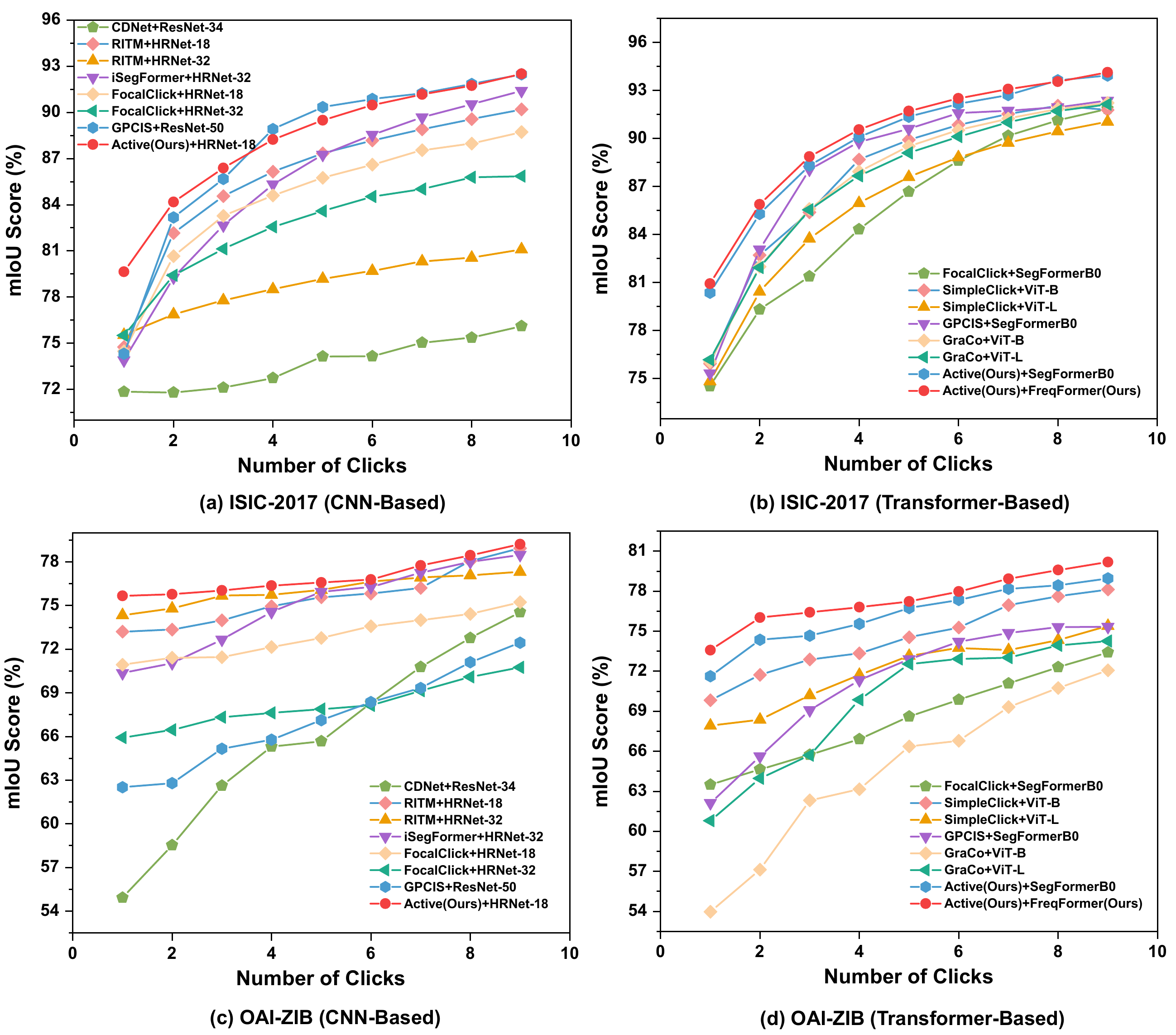}
	\caption{mIoU scores with varying numbers of user clicks on two medical datasets. (a) and (b) are CNN-based methods and Transformer-based methods on ISIC-2017~\cite{berseth2017isic}. (c) and (d) show CNN-based and Transformer-based models on OAI-ZIB~\cite{ambellan2019oaizib}.}
	\label{FIG:6}
\end{figure*}

Fig.~\ref{FIG:6} offers a comprehensive analysis of mIoU performance across varying numbers of user clicks, highlighting the convergence efficiency of different methods. On both ISIC-2017\cite{berseth2017isic} and OAI-ZIB~\cite{ambellan2019oaizib} datasets, ActiveFreq consistently requires fewer clicks to achieve comparable or superior performance compared to competing approaches. For instance, on ISIC-2017, ActiveFreq surpasses 90\% mIoU with only four clicks, while methods such as SimpleClick~\cite{liu2023simpleclick} and FocalClick~\cite{chen2022focalclick} need eight or more clicks to reach similar levels. Likewise, on OAI-ZIB, ActiveFreq achieves 78\% mIoU with just six clicks, outperforming other methods that require additional user interactions to reach comparable performance. These findings highlight the rapid convergence of ActiveFreq, making it a robust and efficient choice for interactive segmentation tasks.

\noindent \textbf{Computational Analysis.} Tab.~\ref{table5} shows a comparison of computational requirements with respect to model parameters, FLOPs, and speed. For fair comparison, we evaluate all the models on the same benchmark (\emph{i.e.} FocalClick) and using the same computer (GPU: NVIDIA RTX4090, CPU: Intel(R) Xeon(R) Silver). By default, our method takes images of size 448$\times$448 as the fixed input. According to the table, we observe that our proposed FreqFormer achieves the best performance while maintaining a lightweight architecture with a minimal number of parameters (9.20 M) and computational complexity (2.67 G).

\begin{table}[htbp]
    \caption{Computation comparison for model parameters, FLOPs, and speed (measured by seconds per click). The best scores are highlighted in bold, and the second-best are underlined.}
    \label{table5}
    \centering
    \renewcommand\arraystretch{1.2}
    \resizebox{0.5\textwidth}{!}{
    \begin{tabular}{lccc}
        \Xhline{1px}
            Backbone & Params/M & FLOPs/G & SPC/ms \\ \hline
            ResNet34~\cite{he2016deep} & 23.47 & 113.60 & \textbf{34} \\
            Resnet50~\cite{he2016deep} & 40.36 & 78.82 & 331 \\
            HRNet18~\cite{sun2019high} & 10.03 & 30.99 & 56 \\
            HRNet32~\cite{sun2019high} & 30.95 & 83.12 & 86 \\
            Swin-Base~\cite{liu2021swin} & 87.44 & 138.21 & \underline{36} \\
            Swin-Large~\cite{liu2021swin} & 195.90 & 302.78 & 44 \\
            ViT-Base~\cite{dosovitskiy2020image} & 96.46 & 42.44 & 54 \\
            ViT-Large~\cite{dosovitskiy2020image} & 322.18 & 133.22 & 86 \\
            SegFormerB0~\cite{xie2021segformer} & \textbf{3.72} & \textbf{1.75} & 37 \\
            SegFormerB3~\cite{xie2021segformer} & 45.66 & 24.75 & 53 \\
            FreqFormer (Ours) & \underline{9.20} & \underline{2.67} & 46 \\
        \Xhline{1px}
    \end{tabular}
    }
\end{table}

\begin{figure*}
	\centering
	\includegraphics[scale=0.47]{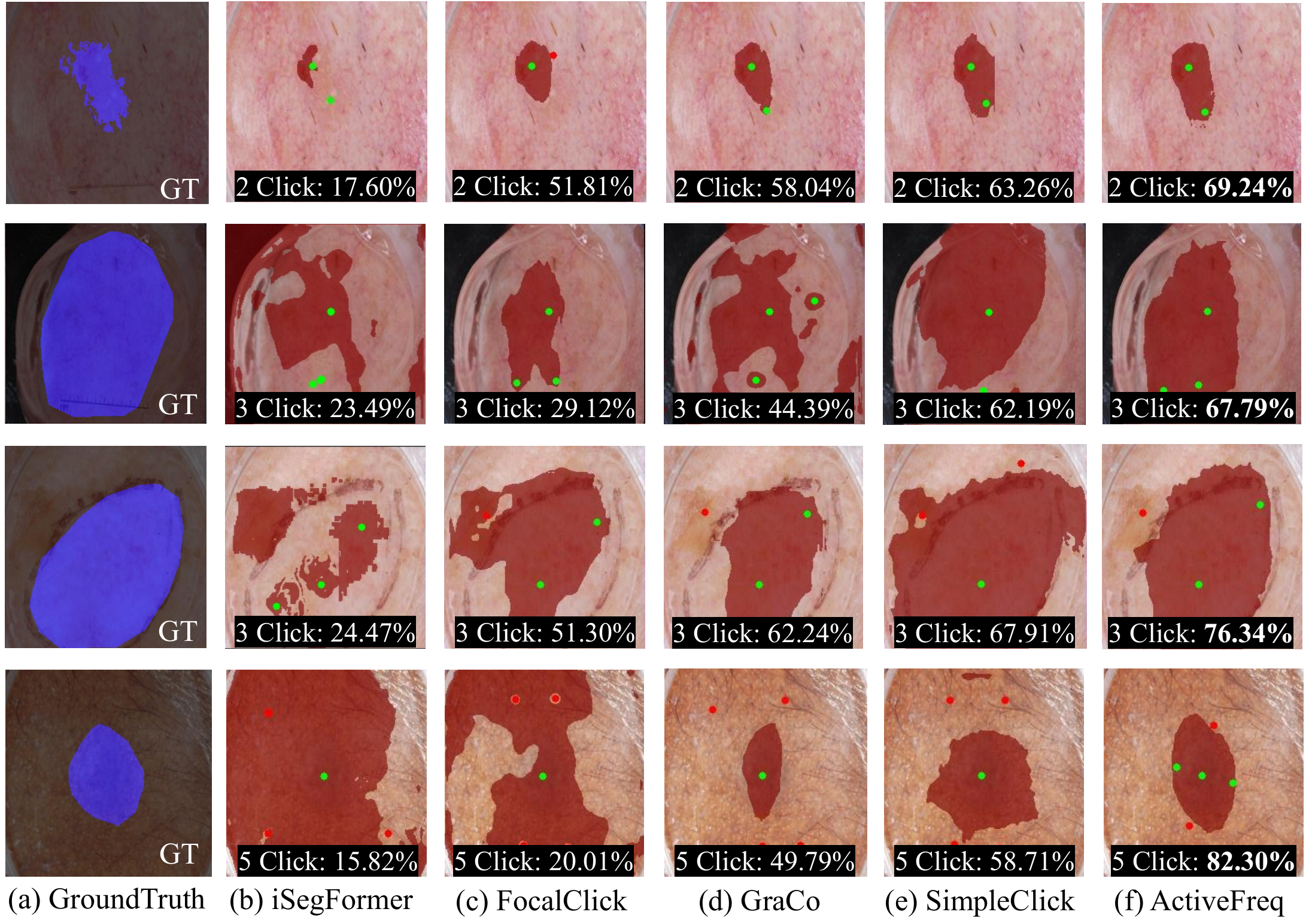}
	\caption{Qualitative results on ISIC-2017~\cite{berseth2017isic}. Each row shows (a) ground truth, and segmentation results for (b) iSegFormer~\cite{liu2022isegformer} using Swin-B, (c) FocalClick~\cite{chen2022focalclick} with SegFormerB0, (d) GraCo~\cite{Zhao_2024_CVPR} based on ViT-B, (e) SimpleClick~\cite{liu2023simpleclick} with SegFormerB0, (f) the proposed ActiveFreq. Green points indicate positive clicks and red points indicate negative clicks. The red masks represent segmentation results, with numerical values indicating the Intersection over Union (IoU) between the predicted mask and the ground truth.}
	\label{FIG:7}
\end{figure*}

\subsubsection{Visualization Analysis}
Fig.~\ref{FIG:7} presents qualitative comparisons of segmentation results on the ISIC-2017\cite{berseth2017isic} dataset, highlighting both the segmentation performance and the effectiveness of AcSelect in determining optimal click regions.

The green points indicate positive clicks, while the red points denote negative clicks. Unlike other methods that often randomly select a mislabeled region and place clicks arbitrarily, ActiveFreq, guided by AcSelect, prioritizes the most informative mislabeled regions for user interactions. This targeted selection ensures that each click contributes maximally to refining segmentation quality. It can be seen that clicks of ActiveFreq are strategically placed near regions of high uncertainty or along critical edges, resulting in more precise and complete segmentations. In contrast, GraCo~\cite{Zhao_2024_CVPR} and SimpleClick~\cite{liu2023simpleclick} often distribute clicks in less impactful regions, leading to fragmented or incomplete masks.

In the last row of Fig.~\ref{FIG:7}, ActiveFreq demonstrates its ability to handle complex boundaries by aligning closely with the ground truth after just five clicks, achieving an IoU of 82.30\%. iSegFormer~\cite{liu2022isegformer}, FocalClick~\cite{chen2022focalclick}, and SimpleClick~\cite{liu2023simpleclick} show many False Positive regions, as their click placement fails to adequately address key segmentation errors. This highlights the advantage of AcSelect’s active learning approach in optimizing click allocation.

\begin{figure*}
	\centering
	\includegraphics[scale=0.47]{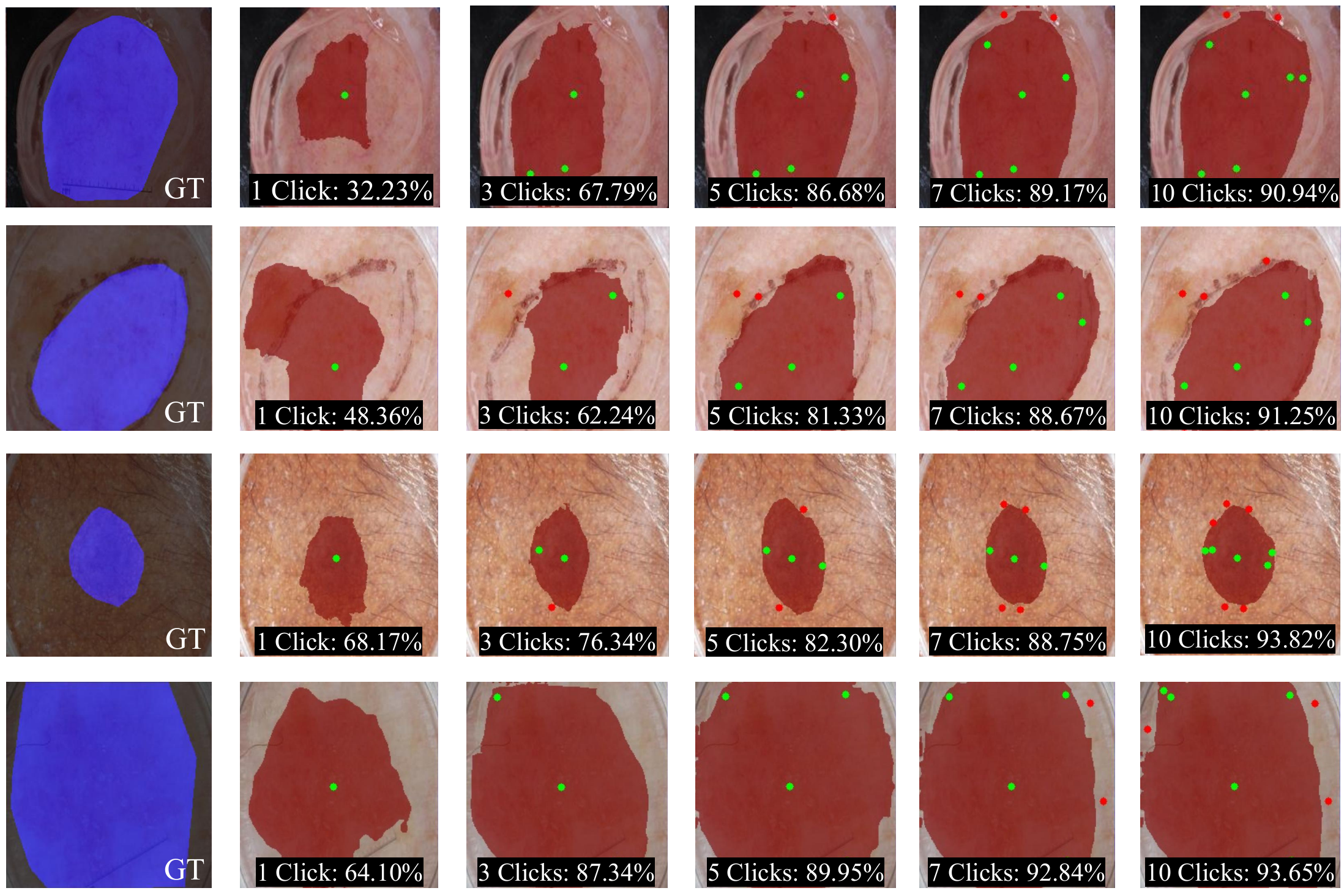}
	\caption{Visualization of the mIoU performance for interactive segmentation on four different medical samples using our proposed ActiveFreq with  varying click counts. The results demonstrate how the performance improves as the number of clicks increases from 1 to 10.}
	\label{FIG:8}
\end{figure*}

In addition, Fig.~\ref{FIG:8} presents the results of the ActiveFreq on the ISIC-2017~\cite{berseth2017isic} dataset, illustrating how incremental user clicks improve performance on four distinct samples. As the number of clicks increases from 1 to 10, the mIoU score significantly improves across all samples, highlighting the model's ability to refine segmentation with few user interaction. Notably, ActiveFreq achieves a near 90\% mIoU score by the seventh click, demonstrating the superiority and effectiveness of the AcSelect strategy in selecting the most informative mislabeled regions.

\subsection{Ablation Study}
We conduct ablation studies to verify the effectiveness of key components in AcSelect and FreqModule on the ISIC-2017~\cite{berseth2017isic} and OAI-ZIB~\cite{ambellan2019oaizib} datasets.

\subsubsection{AcSelect} 
Tab.~\ref{table2} presents the ablation results for AcSelect, where we assess the contributions of its three key metrics: maximum pixel entropy (MPE), average pixel entropy (APE), and regional group uncertainty (RGU), by progressively incorporating them into the model.

\begin{table}[htbp]
    \caption{Ablation study results of AcSelect. We use ActiveFreq (AcSelect with FreqFormer) trained on the ISIC-2017~\cite{berseth2017isic} and OAI-ZIB~\cite{ambellan2019oaizib} datasets. The experiments incrementally added the three key metrics (MPE, APE, and RGU) to assess their individual and combined contributions to segmentation performance. Results are reported in terms of NoC@85 and NoC@90.}
    \label{table2}
    \centering
    \resizebox{0.5\textwidth}{!}{
    \begin{tabular}{ccc|cc|cc}
        \Xhline{1px}
        \multirow{2}{*}{MPE} & \multirow{2}{*}{APE} & \multirow{2}{*}{RGU} & \multicolumn{2}{c|}{ISIC-2017~\cite{berseth2017isic}} & \multicolumn{2}{c}{OAI-ZIB~\cite{ambellan2019oaizib}} \\
         & & & NoC@85 & NoC@90 & NoC@85  & NoC@90 \\ 
        \hline
         & & & 3.03 & 5.11 & 8.31 & 9.62 \\
        \checkmark & & & 2.75 & 4.62 & 7.83 & 9.49 \\
        \checkmark & \checkmark & & 2.43 & 4.19 & 7.37 & 9.37\\
        \checkmark & \checkmark & \checkmark & \textbf{2.21} & \textbf{3.74} & \textbf{6.91} & \textbf{9.27} \\ 
        \Xhline{1px}
    \end{tabular}
    }
\end{table}

Without any metrics, the NoC@90 on ISIC-2017~\cite{berseth2017isic} and OAI-ZIB~\cite{ambellan2019oaizib} are 5.11 and 9.62, respectively. Adding MPE reduces NoC@90 to 4.62 and 9.49, confirming its effectiveness in targeting the most uncertain pixels and refining localized errors. Including APE further lowers NoC@90 to 4.19 and 9.37, demonstrating its ability to capture overall uncertainty and balance the model’s focus. Finally, adding RGU achieves the best results, with NoC@90 decreasing to 3.74 and 9.27, showcasing its capacity to enhance regional consistency and refine uniformly uncertain areas.

From these improvements, MPE provides the most significant initial gain by addressing localized uncertainty, making it essential for immediate refinement. APE complements MPE by accounting for broader uncertainty across regions, contributing further gains. While RGU has a smaller standalone impact, it is crucial when combined with MPE and APE, ensuring consistent and robust predictions. These results demonstrate that each metric contributes uniquely and complements the others in enhancing segmentation performance.

Furthermore, to demonstrate the effectiveness of AcSelect’s sampling strategy, we conducted a comparative analysis against Random Sampling, Entropy-Based Sampling, and Least Confidence Sampling. Tab.~\ref{table7} presents the results of the sampling strategy ablation study, comparing Random Sampling, Entropy-Based Sampling, Least Confidence Sampling, and our proposed AcSelect method on the ISIC-2017 and OAI-ZIB datasets. The results indicate that AcSelect consistently achieves the best performance across both datasets, demonstrating its superior ability to select the most informative samples for annotation. Notably, compared to Random Sampling, AcSelect reduces NoC@85 by 27.06\% on ISIC-2017 and 16.83\% on OAI-ZIB, highlighting its efficiency in minimizing the number of clicks required to reach the convergence threshold. Although entropy and least confidence improve performance in random sampling, AcSelect significantly outperforms them, further validating its effectiveness in enhancing interactive segmentation through optimized sample selection.

\begin{table}[htbp]
    \caption{Ablation study results on the sampling strategy. This table compares different active learning sampling strategies, including Random Sampling, Entropy-Based Sampling, Least Confidence Sampling. The best results are highlighted in bold.}
    \label{table7}
    \centering
    \renewcommand\arraystretch{1.2}
    \resizebox{0.5\textwidth}{!}{
    \begin{tabular}{l|cc|cc}
        \Xhline{1px}
        \multirow{2}{*}{Sampling Strategy} & \multicolumn{2}{c|}{ISIC-2017} & \multicolumn{2}{c}{OAI-ZIB} \\
        & NoC@85 & NoC@90 & NoC@85 & NoC@90 \\ \hline
        Random & 3.03 & 5.11 & 8.31 & 9.62 \\
        Entropy & 2.71 & 4.68 & 7.85 & 9.50 \\
        Least Confidence & 2.69 & 4.45 & 7.72 & 9.45 \\
        AcSelect (Ours) & \textbf{2.21} & \textbf{3.74} & \textbf{6.91} & \textbf{9.27} \\
        \Xhline{1px}
    \end{tabular}
    }
\end{table}

\begin{table}[htbp]
    \caption{Ablation study results on the FreqModule. We use ActiveFreq (AcSelect with FreqFormer) trained on ISIC-2017~\cite{berseth2017isic} and OAI-ZIB~\cite{ambellan2019oaizib}. Experiments include progressively adding 2D DFT transformations ($F_{(H, W)}$, $F_{(H, C)}$, and $F_{(W, C)}$) to evaluate their effectiveness.}
    \label{table3}
    \centering
    \resizebox{0.5\textwidth}{!}{
    \begin{tabular}{ccc|cc|cc}
        \Xhline{1px}
        \multirow{2}{*}{$F_{(H,W)}$} & \multirow{2}{*}{$F_{(H,C)}$} & \multirow{2}{*}{$F_{(W,C)}$} & \multicolumn{2}{c|}{ISIC-2017~\cite{berseth2017isic}} & \multicolumn{2}{c}{OAI-ZIB~\cite{ambellan2019oaizib}} \\
         & & & NoC@85 & NoC@90 & NoC@85 & NoC@90 \\ 
        \hline
         & & & 3.14 & 5.22 & 8.42 & 9.73 \\
        \checkmark & & & 2.85 & 4.74 & 7.94 & 9.61 \\
        \checkmark & \checkmark & & 2.52 & 4.28 & 7.48 & 9.48 \\
        \checkmark & \checkmark & \checkmark & \textbf{2.21} & \textbf{3.74} & \textbf{6.91} & \textbf{9.27} \\ 
        \Xhline{1px}
    \end{tabular}
    }
\end{table}

\subsubsection{FreqModule}  
Tab.~\ref{table3} presents the ablation results for the FreqModule, where 2D DFT transformations (\( F_{(H, W)} \), \( F_{(H, C)} \), and \( F_{(W, C)} \)) are progressively added to evaluate their contributions. Without any transformations, the NoC@90 values on ISIC-2017~\cite{berseth2017isic} and OAI-ZIB~\cite{ambellan2019oaizib} are 5.22 and 9.73, respectively. Each transformation individually improves performance, with all three combined achieving the best results: reducing NoC@90 to 3.74 on ISIC-2017~\cite{berseth2017isic} and 9.27 on OAI-ZIB~\cite{ambellan2019oaizib}.

The results indicate that each transformation plays a crucial role in improving segmentation precision by capturing complementary information. The \( F_{(H, W)} \) transformation filters high-frequency noise while preserving low-frequency structural patterns, essential for delineating global segmentation boundaries. The \( F_{(H, C)} \) and \( F_{(W, C)} \) transformations enrich contextual understanding by capturing correlations between feature channels and spatial dimensions, which highlight variations in texture and intensity across regions. The similar magnitude of improvement across all dimensions underscores their complementary roles, collectively enabling the FreqModule to integrate multi-dimensional frequency information effectively. This integration enhances feature representation, reduces user interactions, and leads to robust segmentation performance.

\begin{table}[htbp]
    \caption{Ablation study results on the Frequency Module. This table presents a comparison between Fourier Transform (FT) and Wavelet Transform (WT) in the frequency module across the ISIC-2017 and OAI-ZIB datasets. While WT demonstrates some capability in frequency domain representation, FT consistently outperforms WT across all metrics. The better results are highlighted in bold.}
    \label{table6}
    \centering
    \renewcommand\arraystretch{1.2}
    \resizebox{0.5\textwidth}{!}{
    \begin{tabular}{c|cc|cc}
        \Xhline{1px}
        \multirow{2}{*}{Frequency Module} & \multicolumn{2}{c|}{ISIC-2017} & \multicolumn{2}{c}{OAI-ZIB} \\
        & NoC@85 & NoC@90 & NoC@85 & NoC@90 \\ \hline
        Wavelet Transform & 2.66 & 3.83 & 7.09 & 9.40 \\
        Fourier Transform & \textbf{2.21} & \textbf{3.74} & \textbf{6.91} & \textbf{9.27} \\
        \Xhline{1px}
    \end{tabular}
    }
\end{table}

\subsubsection{Wavelet Transform vs. Fourier Transform}
To further investigate the impact of frequency domain transformations on segmentation performance, we conducted an ablation study by replacing the Fourier Transform (FT) with the Wavelet Transform (WT) in our method. Tab.~\ref{table6} indicates that although WT enables multi-scale analysis by decomposing features into different frequency bands with spatial localization, FT outperforms WT in capturing global contextual information, leading to better overall segmentation performance. Although WT shows some advantages in fine-structure segmentation and complex textures, it falls short in maintaining global consistency. Based on these findings, we retain FT as the frequency transformation in our approach.

\section{Limitations}  
While ActiveFreq demonstrates significant advancements in interactive segmentation, some limitations remain. 

First, AcSelect analyzes mislabeled regions individually, potentially missing interactions between regions. The grouping and collectively analyzing regions could improve efficiency by focusing on areas that contribute most to global segmentation accuracy. Second, the FreqModule applies 2D Discrete Fourier Transform (2D DFT) operations across height-width, height-channel, and width-channel dimensions, capturing pairwise frequency information but missing interactions across all dimensions. Future work could explore 3D Discrete Fourier transform (3D DFT) for holistic frequency feature extraction and investigate adaptive weighting or learnable filters to better integrate frequency features and improve segmentation performance.

\section{Conclusions}  
In this work, we propose \textbf{ActiveFreq}, an interactive image segmentation framework that reduces human intervention while maintaining high segmentation performance. ActiveFreq addresses two key limitations of existing methods: the uniform treatment of all mislabeled regions and the lack of frequency domain analysis. To overcome these challenges, we introduce \textbf{AcSelect}, an active learning selection module that prioritizes the most informative mislabeled regions, optimizing refinement efficiency and minimizing user interactions. Additionally, we develop \textbf{FreqFormer}, a segmentation network that integrates spatial and frequency domain features using Fourier transform modules, enhancing feature representation and segmentation accuracy. Extensive experiments on the ISIC-2017~\cite{berseth2017isic} and OAI-ZIB~\cite{ambellan2019oaizib} datasets validate the effectiveness of ActiveFreq. Our method achieves state-of-the-art performance, with NoC@90 values of 3.74 on ISIC-2017 and 9.27 on OAI-ZIB. These results demonstrate the great potential of ActiveFreq to advance interactive segmentation and contribute to the development of medical image analysis algorithms.

\printcredits

\section*{Declaration of competing interest}
The authors declare that they have no competing financial interests or personal relationships that could influence the work reported in this paper.

\section*{Acknowledgments}
This work was supported in part by Key Research and Development Program of Hubei Province (2022BCA009); Bingtuan Science and Technology Program (2022DB005); Dongguan Science and Technology of Social Development Program (20231800935472). (Corresponding author: Hua Zou.)

\section*{Data availability}
The study used publicly available datasets, including ISIC-2017 (available at \url{https://challenge.isic-archive.com/data/\#2017}) and OAI-ZIB (available at \url{https://drive.google.com/drive/folders/1B6y1nNBnWU09EhxvjaTdp1XGjc1T6wUk}).


\bibliography{cas-refs}

\end{document}